\documentclass[10pt,twocolumn,letterpaper]{article}

\usepackage[pagenumbers]{iccv} 
\usepackage{times}
\usepackage{epsfig}
\usepackage{graphicx}
\usepackage{amsmath}
\usepackage{amssymb}
\usepackage{xspace}
\usepackage{multirow}
\usepackage{xcolor}
\usepackage{makecell,rotating}
\usepackage{booktabs}
\usepackage{colortbl} 
\usepackage{subcaption}
\usepackage[margin=1in]{geometry}
\definecolor{iccvblue}{rgb}{0.21,0.49,0.74}
\usepackage[pagebackref,breaklinks,colorlinks,allcolors=iccvblue]{hyperref}

\def\paperID{5175} 
\def\confName{ICCV}
\def\confYear{2025}

\title{Leveraging Textual Anatomical Knowledge for Class-Imbalanced Semi-Supervised Multi-Organ Segmentation}

\author{Yuliang Gu$^{1}$$^{\dagger}$ , Weilun Tsao$^{1}$$^{\dagger}$, Bo Du$^{1}$, Thierry G\'{e}raud$^{2}$, Yongchao Xu$^{1}$
\thanks{Corresponding author}
\\
1 School of Computer Science, Wuhan University, Wuhan 430072, China\qquad\\
2 EPITA Research Laboratory, Le Kremlin-Bic\^etre, France\qquad
\\
{\tt\small \{yuliang\_gu, dubo, yongchao.xu\}@whu.edu.cn,\, \{wiwilliam8855\}@gmail.com}
\\
{\tt\small \{thierry.geraud\}@epita.fr}
}

\begin{document}
\maketitle
\begin{abstract}
Imbalanced class distributions among different organs pose significant challenges in real-world semi-supervised multi-organ segmentation.
Integrating anatomical priors offers a promising research direction to mitigate these imbalances.
In this paper, we exploit the capabilities of Multimodal Large Language Models (MLLM) to extract robust, generic textual anatomical insights serving as prior knowledge for segmentation model. Specifically, we employ GPT-4o to generate detailed textual descriptions of anatomical priors—including both inter-organ spatial relationships and organ shape characteristics. These priors are then seamlessly integrated into the segmentation model as parameters within the segmentation head. Furthermore, we align the textual priors with visual features using contrastive learning. The inter-organ positional priors guide the model in localizing smaller organs relative to larger ones, while the organ shape priors help ensure that the learned morphological structures are more anatomically plausible.
Extensive experiments demonstrate that our method significantly outperforms current state-of-the-art approaches. The source code is available at: https://github.com/Lunn88/TAK-Semi.
\end{abstract}    
\section{Introduction}
\label{sec:intro}

\begin{figure}[t]
\centering
\includegraphics[width = 1\linewidth]{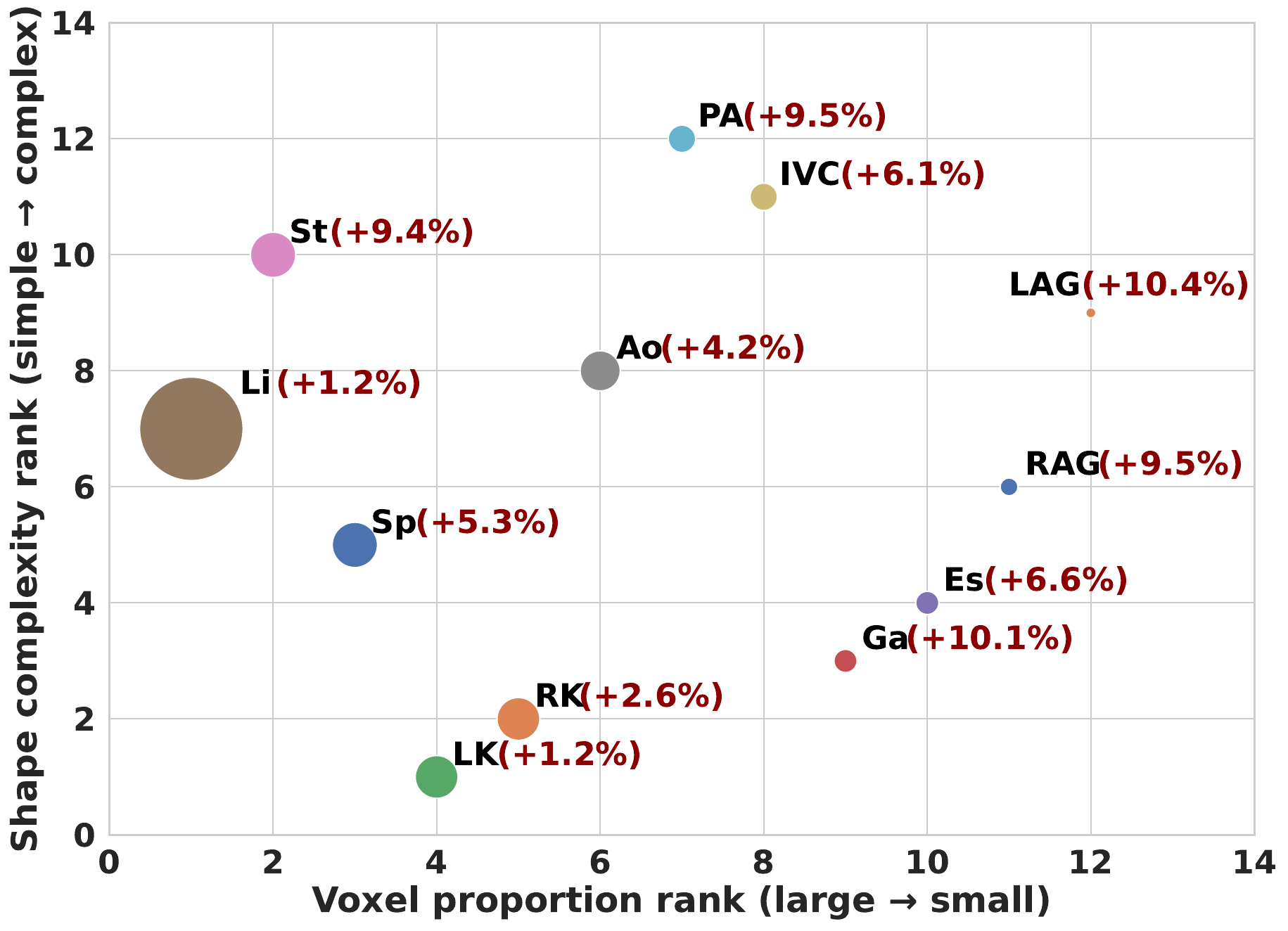}
\caption{A scatter plot of the voxel proportion and shape complexity of different organs. The shape complexity is represented by Convex Hull Volume Ratio. {Red numbers} in parentheses represent average improvement of Dice score compared with state-of-the-art method GA~\cite{wu98gradient}. The size of each point represents the proportion of their voxel volume. 
} 
\label{fig:shape_complex}
\end{figure}

Accurate multi-organ segmentation is crucial for computer-aided diagnosis (CAD). While supervised methods perform well with large labeled datasets, manual annotation is labor-intensive. Semi-supervised segmentation has gained attention by leveraging unlabeled images to improve accuracy~\cite{chen2022semi}.
These approaches generally rely on two key strategies: consistency regularization~\cite{ basak2023pseudo,xiang2022fussnet, gu2025dual} and pseudo-labeling~\cite{chen2021semi,lyu2022pseudo,qiao2022semi}. Consistency regularization is based on the assumption that model predictions should remain stable under small data or/and model perturbations, promoting smoother and more reliable results~\cite{chen2022semi}. Pseudo-labeling~\cite{chen2021semi,lyu2022pseudo,qiao2022semi}, in contrast, leverages model-generated predictions on unlabeled data to create pseudo-labels, effectively augmenting the initially limited labeled dataset.

The intricate complexity of human anatomy results in significant variations in the voxel proportions of different organs in medical images~\cite{wang2023dhc}. Larger organs, such as the liver and stomach, occupy substantial portions of the voxel space, while smaller organs, like the pancreas and esophagus, account for only a fraction.
This imbalance complicates the model's ability to learn and maintain balanced feature representations across categories~\cite{wang2023dhc}. 
For organs with smaller voxel proportions, models often struggle to capture their features effectively, resulting in lower segmentation accuracy.
The challenge is further exacerbated in semi-supervised settings, where the limited availability of labeled data for smaller organs further influences training and causes noticeable declines in segmentation accuracy.
Recently, several approaches~\cite{wang2023dhc,lin2022cld,chen2023magicnet,wu98gradient} have been proposed to address the challenge of class imbalance in semi-supervised medical image segmentation. 
These methods address class imbalance in segmentation by adjusting loss functions~\cite{lin2022cld}, incorporating class distribution and learning difficulty~\cite{wang2023dhc}, using data augmentation~\cite{chen2023magicnet, gu2024shape}, and mitigating gradient bias~\cite{wu98gradient}.

The anatomical structure of the multi-organ in human body holds valuable prior information, such as inter-organ spatial relationships and organ shape priors. 
For instance, the \textbf{duodenum} typically exhibits fairly consistent positional relationships with adjacent organs such as the \textbf{pancreas}, \textbf{stomach}, \textbf{liver}, \textbf{gallbladder}, and \textbf{kidneys}. As illustrated in Fig.~\ref{fig:motivation} (a) - Fig.~\ref{fig:motivation} (d), especially in Fig.~\ref{fig:motivation} (b), \textbf{\textit{the duodenum lies posterior to the stomach and anterior to the inferior vena cava and aorta.}} Due to its relatively small voxel proportion, segmentation models frequently overlook the duodenum. Consequently, a key question is how to leverage its inter-organ spatial relationships (especially its relationships with larger organs, like stomach, kidney, etc.) to improve segmentation performance for this challenging category.
Furthermore, as noted in~\cite{wang2023dhc}, in class-imbalanced semi-supervised multi-organ segmentation tasks, the model not only yields poor segmentation performance for categories with a small voxel proportion, but also struggles with organs that have larger voxel proportions yet complex morphology, such as the \textbf{stomach}.
Another question is how to flexibly and robustly inject morphology and shape prior into the model to improve the segmentation accuracy for categories with complex morphology. 

In this paper, we propose a novel semi-supervised framework that integrates anatomical prior knowledge derived from Multimodal Large Language Models (MLLMs) to address class imbalance in multi-organ segmentation. Our approach is driven by two key insights:
(1) Inter-organ Spatial Guidance: Leveraging inter-organ spatial relationship priors to guide the localization of smaller organs. (2) Shape-Aware Regularization: Encoding organ shape priors to constrain segmentation outputs to more anatomically plausible forms, particularly for organs with complex geometries. Specifically, we employ GPT-4o to generate structured textual descriptions of anatomical relationships and morphological patterns, transforming implicit domain knowledge into explicit, model-actionable priors. These priors are encoded as adaptive parameters within the segmentation head, enabling context-aware feature refinement during both supervised and unsupervised training phases. Furthermore, we design cross-modal contrastive alignment module to align visual features with textual priors in a shared embedding space, ensuring consistency between pixel-level predictions and anatomical constraints.
As shown in Fig.~\ref{fig:shape_complex}, we use the convex hull volume ratio to quantify the complexity of organ shapes and present a scatter plot illustrating the relationship between organ complexity and the voxel proportion they occupy. The convex hull volume ratio is defined as the ratio between the volume of a geometric shape's convex hull and its original volume. The numbers in parentheses indicate the performance improvement of our method over the GA~\cite{wu98gradient} method.
Our method significantly boosts segmentation accuracy for challenging categories. 

The key contributions can be summarized as follows:
 \textbullet~We propose the framework to leverage MLLMs for generating robust, generic textual anatomical priors, including inter-organ spatial relationships and organ shape characteristics. MLLMs transform implicit knowledge into explicit, model-actionable priors.

 \textbullet~We encode these priors as adaptive parameters in the segmentation head and design the cross-modal contrastive alignment to ensure that predictions adhere to both visual evidence and anatomical plausibility.

 \textbullet~Beyond demonstrating significant improvements across various multi-organ datasets, we contribute to the research community by making the codebase publicly available, facilitating reproducibility and further exploration in related tasks.
\section{Related Work}

\subsection{Class-imbalanced semi-supervised medical image segmentation}
Semi-supervised learning is widely used in medical image segmentation to reduce manual annotation efforts. Consistency regularization methods~\cite{basak2023pseudo, xiang2022fussnet}, such as Mean Teacher~\cite{tarvainen2017mean} and Co-training~\cite{qiao2018deepcotraining}, enhance segmentation by introducing model-level variations. Additionally, an increasing number of techniques enhance model performance by leveraging pseudo labels for training on unlabeled images~\cite{lyu2022pseudo, qiao2022semi}. Given the inevitable presence of noisy labels in pseudo labels for unlabeled images, many methods~\cite{qiao2022semi, wang2021semi} select pseudo labels with high confidence as the labels for unlabeled data based on their confidence levels. 
Recently, many studies have demonstrated the effectiveness of contrastive learning to enhance the representation capability of models for unlabeled data in the field of semi-supervised segmentation~\cite{basak2023pseudo,wang2023hunting,alonso2021semi}.

Real-world datasets often suffer from class imbalances, complicating machine learning model training and generalization~\cite{liu2019large}. To address this, techniques such as re-weighting~\cite{cao2019learning,shu2019meta} and re-sampling~\cite{byrd2019effect} are commonly used. Re-weighting adjusts the loss function to give more weight to minority class samples, while re-sampling changes the label distribution by over-sampling the minority class or under-sampling the majority class. Due to the limited labeled data in semi-supervised learning, the class imbalance problem makes it more difficult for the model to learn from underrepresented classes, posing a significant challenge for extending existing SSL-based methods to more practical settings. CReST~\cite{wei2021crest} uses self-training to select pseudo-labeled data to balance the class distribution, favoring minority classes. 
CLD~\cite{lin2022cld} adjusts the loss by weighting classes based on their instance count. DHC~\cite{wang2023dhc} further mitigates data and learning biases by considering both class distribution and learning difficulty. MagicNet~\cite{chen2023magicnet} introduces a partition-and-recovery $N^{3}$ cube augmentation strategy for better learning of small organs. Qi~\etal~\cite{wu98gradient} identify gradient biases in class-imbalanced semi-supervised segmentation and propose Gradient-Aware loss to address these issues.

\subsection{CLIP-based medical image analysis}
Large-scale vision-language models~\cite{radford2021clip,liu2024visual,li2022grounded} (VLMs) integrate information from both text and image modalities, enhancing their ability to understand and generate cross-domain knowledge, thereby improving performance and generalization in complex tasks. CLIP~\cite{radford2021clip} has recently gained popularity in medical imaging, serving as a pre-training method for image-text alignment and playing a crucial role in downstream tasks~\cite{zhang2023biomedclip, khattak2024unimed, lin2023pmc}. 
 BiomedCLIP~\cite{zhang2023biomedclip}, a multimodal foundation model for biomedical vision-language processing, was pretrained on the PMC-15M dataset, which consists of 15 million biomedical image-text pairs. 
UniMed-CLIP~\cite{khattak2024unimed}, a unified vision-language model trained on the large-scale, open-source multi-modal medical dataset UniMed, leverages over 5.3 million image-text pairs across six imaging modalities for enhanced multi-modal medical task performance.
 Liu~\etal~\cite{liu2023clip} propose the CLIP-Driven Universal Model, using CLIP-based text embeddings for segmentation. Unlike their approach, which relies only on class names, our method integrates textual anatomical knowledge to address class imbalance in semi-supervised learning.
Following Liu~\etal~\cite{liu2023clip}, Zhang~\etal~\cite{zhang2023continual} modify this framework for continual learning, employing extra heads and text prompts.


\subsection{Incorporating anatomical knowledge in medical image analysis}
Organs in medical images contain valuable prior information, and effectively utilizing this information is a key characteristic that distinguishes medical image analysis from natural image analysis. Previous works~\cite{bloch2003representation} utilize fuzzy spatial representations for relative positioning. CAT~\cite{huang2024cat} uses 3D cropped images as anatomical prompts and introduces ShareRefiner to coordinate text descriptions with visual anatomical structures. 
Unlike CAT~\cite{huang2024cat} injecting anatomical priors through visual modules, our approach integrates anatomical knowledge through text descriptions.
Zept~\cite{jiang2024zept} align high-level textual knowledge semantics and visual features to enhance the model's generalization for recognizing unseen tumors. Unlike Zept~\cite{jiang2024zept}, which integrates all textual information as a whole, we separately extract inter-organ relative positions and organ shape priors, enabling contrastive learning to better align anatomical text features with visual features.
Some works~\cite{gupta2022learning, gupta2024topology} use the topological priors of organs as anatomical knowledge. For example, Gupta~\etal~\cite{gupta2022learning} encode topological interactions between organs (e.g., containment and exclusion) into network modules. 
On the other hand,  Li~\etal~\cite{li2020shape} and Luo~\etal~\cite{luo2021semi} use signed distance map and level set to describe the anatomical priors of organs, respectively. 
Different from these works, we use text descriptions to capture inter-organ relative positions and organ morphology, making the anatomical representation richer and more flexible.

\begin{figure}[tb]
    \centering
    \begin{subfigure}[b]{1\linewidth}
        \centering
        \includegraphics[width=\linewidth]{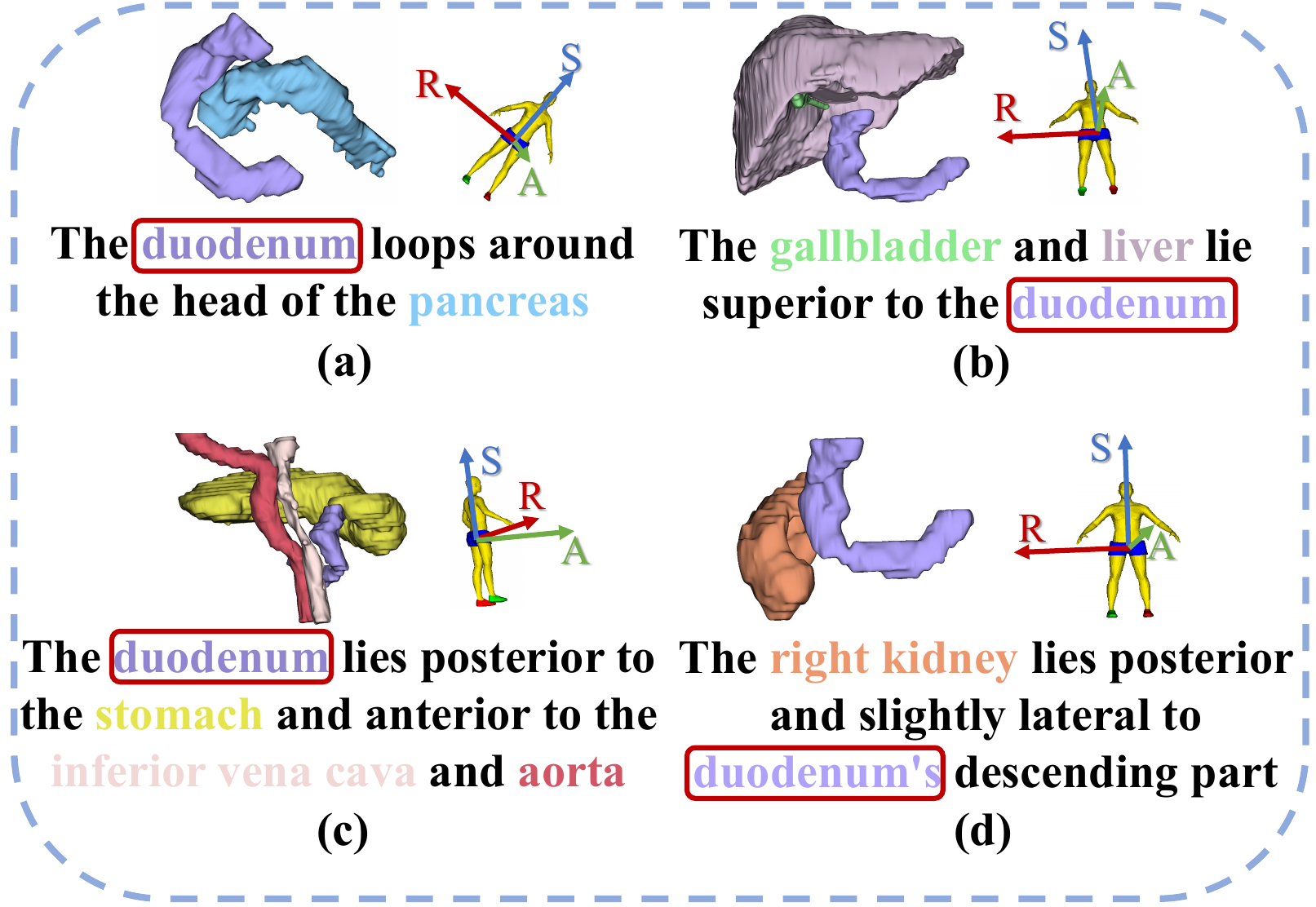}
    \end{subfigure}\hspace{-0mm}

    \begin{subfigure}[b]{1\linewidth}
        \centering
        \includegraphics[width=\linewidth]{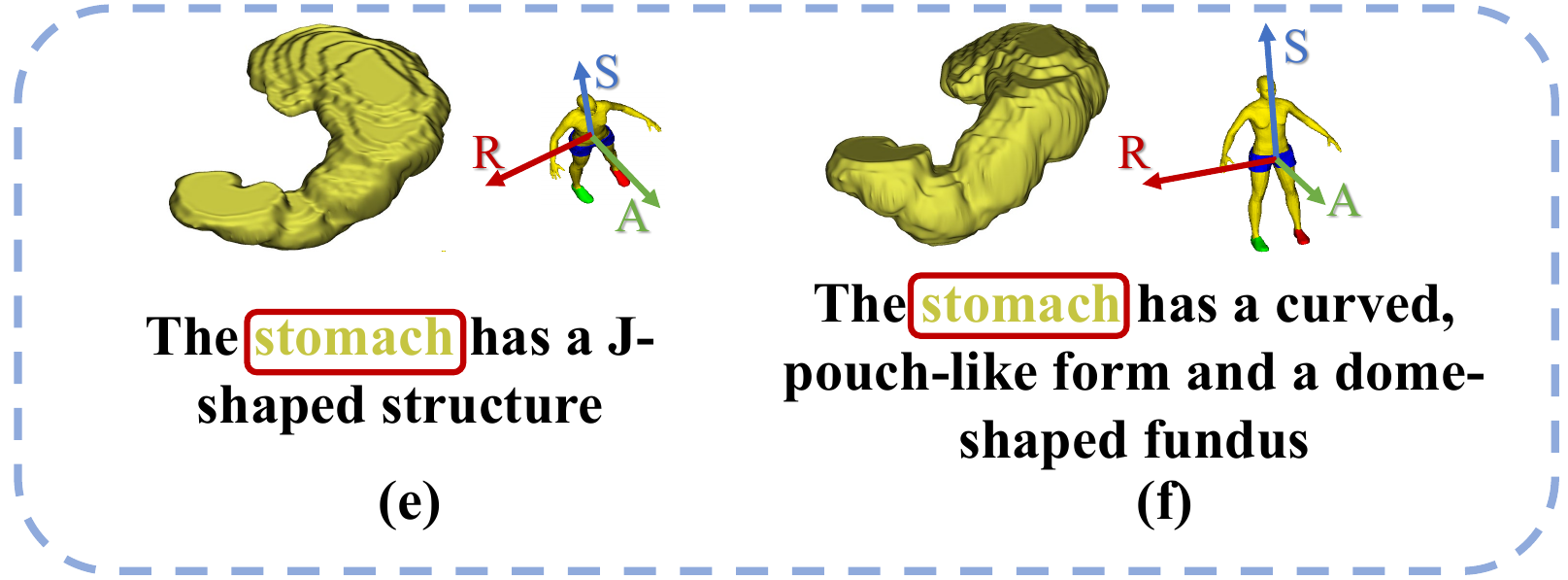}
    \end{subfigure}\hspace{-0mm}

    \caption{Examples of textual anatomical knowledge and their corresponding 3D visualizations. (a)-(d) indicate the inter-organ spatial relationship priors for the duodenum. (e)-(f) indicate the organ shape priors for the stomach. The font color corresponds to the organ in the 3D visualization.}
    \label{fig:motivation}
\end{figure}
\begin{figure*}[t]
\centering
\includegraphics[width = 1\linewidth]{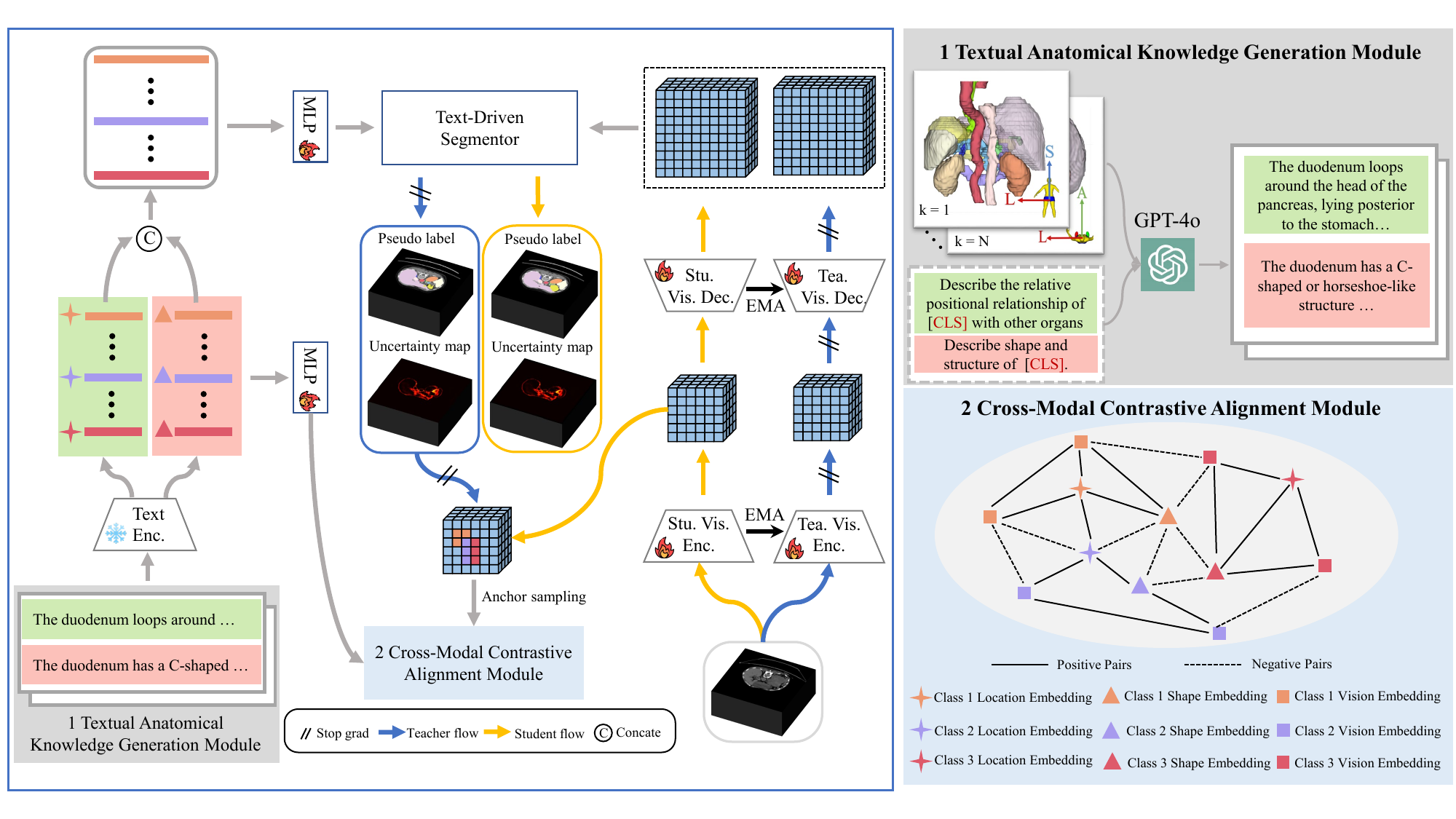}
\caption{Overview of our method. We propose Textual Anatomical Knowledge Generation Module and Cross-Modal Contrastive Alignment Module for generating textual anatomical knowledge and aligning visual features with anatomical prior knowledge, respectively. The blue and orange arrows represent the data flow through the teacher network and student network, respectively. 
It is worth noting that the textual anatomical knowledge only needs to be generated once before the entire training process begins.
} 
\label{fig:pipeline}
\end{figure*}

\section{Method}
\subsection{Overview}
The training set of semi-supervised medical image segmentation includes a limited set of labeled images, $\mathcal{D}^l = \{(x^l, y^l)\}$, containing  $N$ samples, and a significantly larger set of unlabeled images, $\mathcal{D}^u = \{x^u\}$, comprising $M$ samples, where $N\ll M$. 


We propose an anatomy-aware framework that integrates two complementary priors: (1) Inter-organ Spatial Guidance, where leveraging consistent positional relationships between small and adjacent large organs to redirect model attention to underrepresented regions. (2) Shape-Aware Regularization, where encoding organ shape priors to constrain segmentation outputs to more anatomically plausible forms. We first employs GPT-4o to generate textual descriptions of these anatomical priors. Fig.~\ref{fig:motivation} shows some textual descriptions and visualization of these priors. The detail of this Textual Anatomical Knowledge (TAK) Generation process is illustrated in Sec.\ref{sec:llm_generate}. These robust and generic textual anatomical knowledge are then encoded as adaptive parameters in the segmentation head to refine feature decoding. Simultaneously, a contrastive alignment module ensures consistency between visual features and anatomical constraints, bridging language-derived knowledge with pixel-level predictions. This anatomy-guided segmentation process is illustrated in Sec.\ref{sec:anatomical_seg}.  
 The overall framework is illustrated  in Fig.~\ref{fig:pipeline}.

\subsection{MLLM-based textual anatomical knowledge generation.}
\label{sec:llm_generate}
We develop a TAK generation framework using a two-agents system, where each agent is powered by a GPT-4o model. The first agent generates priors on inter-organ relative positions as well as the morphology and shape of organs as broadly as possible. To ensure the generalizability of these priors, we provide input to the first agent solely through text prompts. To generate the $k$-th class inter-organ relative positions knowledge $\mathcal{P}_{k}^{p}$, we first input the names of all categories to be segmented into GPT-4o, then use the template``Describe the relative positional relationship of [CLS] with other organs." On the other hand, for the morphology and shape knowledge $\mathcal{P}_{k}^{s}$, we use ``Describe the shape and structure of [CLS]'' as the template.  To refine these priors, the second agent performs multi-modal validation by cross-checking generated descriptions against visual evidence from randomly sampled labeled data. Additionally, the agent filters low-confidence claims by discarding inconsistent statements. Finally, the generated descriptions are reviewed by professional doctors. For more details about textual anatomical knowledge generation, please see the supplementary.

\subsection{Integrating textual anatomical knowledge into the segmentation framework}
\label{sec:anatomical_seg}
We leverage the text encoder $\textbf{Enc}^{T}$ of BiomedCLIP~\cite{zhang2023biomedclip} to encode the textual anatomical knowledge. The injection of these priors is achieved through two main aspects: text-driven segmentor and cross-modal contrastive alignment module. For the text-driven segmentor, we draw inspiration from the method in ~\cite{liu2023clip} utilizing the text embeddings to produce the segmentation head parameters. For text and visual alignment module, we employ contrastive learning to align the textual features with the visual features.
Let $T_{k}^{p} = \textbf{Enc}^{T}(\mathcal{P}_{k}^{p})$ and $T_{k}^{s} = \textbf{Enc}^{T}(\mathcal{P}_{k}^{s})$ represent the anatomical knowledge embedding of the $k$-th class. 
We begin by concatenating $T_{k}^{p}$, $T_{k}^{s}$, and the global image feature $F$, which is obtained by average pooling the visual features.  Tis combined representation is then fed into a multi-layer perceptron, referred to as the text-based controller, to generate the parameters $\theta_{k}$ of the segmentation head.

\begin{table*}[t]
\scriptsize
\resizebox*{\linewidth}{!}{

\begin{tabular}{c|c|ccc|ccccccccccccccc}
\toprule
\multicolumn{2}{c|}{\multirow{2}{*}{Methods}}  & \multicolumn{3}{c|}{Avg.Dice}  & \multicolumn{15}{c}{Average Dice of Each Class}                                        \\ 
\multicolumn{2}{c|}{}       &All    &\cellcolor{orange!20} L.  &\cellcolor{green!20}S.   & \cellcolor{orange!20}Sp   &\cellcolor{orange!20} RK   &\cellcolor{orange!20} LK   &\cellcolor{green!20} Ga   &\cellcolor{green!20} Es   &\cellcolor{orange!20} Li   & \cellcolor{orange!20}St   &\cellcolor{green!20} Ao   &\cellcolor{green!20} IVC   &\cellcolor{green!20} PA   &\cellcolor{green!20} RAG  &\cellcolor{green!20} LAG  & \cellcolor{green!20}Du   &\cellcolor{orange!20} Bl  &\cellcolor{green!20} P/U  \\\midrule
\multirow{9}{*}{\rotatebox{90}{General}}         & V-Net (fully)   &76.48   &86.63   &69.71  &92.2 &92.2 &93.3 &65.5 &70.3 &95.3 &82.4 &91.4 &85.0 &74.9 &58.6 &58.1 &65.6 &64.4 &58.0
    \\ 
\midrule

& UA-MT~\cite{yu2019uamt}     &33.92   &54.87   &19.97  &62.5 &61.7 &59.8 &17.5 &13.8 &73.4 &39.4 &34.6 &32.4 &26.5 &12.1 &6.5 &15.3 &32.4 &21.0  \\



& CPS~\cite{chen2021semi}      &31.77    &47.70   &21.14   &55.9 &46.9 &53.1 &27.7 &0.0 &66.4 &25.2 &41.8 &45.2 &29.4 &0.1 &0.0 &22.1 &38.7 &24.0 \\

&DeSCO~\cite{cai2023orthogonal} &40.25  &63.98   &24.43 &71.9 &67.4 &70.3 &6.7 &0.0 &73.9 &40.1 &53.7 &56.0 &34.0 &0.0 &0.0 &25.9 &60.3 &43.6 \\

& DePL~\cite{wang2022depl}    &31.50   &49.40   &19.57  &57.1 &49.3 &54.3 &26.6 &0.1 &69.2 &26.2 &41.1 &46.7 &23.9 &0.0 &0.0 &16.7 &40.3 &21.0    \\

&Co-BioNet~\cite{peiris2023uncertainty} &42.82  &58.12   &32.63    &68.0 &55.5 &54.7 &40.5 &32.9 &75.8 &41.8 &56.5 &50.8 &27.5 &0.0 &20.2 &19.1 &52.9 &\textbf{46.2} \\

&MagicNet~\cite{chen2023magicnet} &47.29    &67.67   &33.71  &69.4 &68.4 &70.3 &46.7 &0.0 &82.7 &55.0 &67.3 &63.3 &53.8 &0.0 &0.0 &36.9 &60.2 &35.4\\

\midrule
\multirow{8}{*}{\rotatebox{90}{Imbalance}} 

& Adsh~\cite{guo2022adsh}     &30.26    &45.85   & 19.88  &53.9 &45.1 &51.2 &28.5 &0.0 &62.1 &27.0 &41.4 &42.7 &25.0 &0.0 &0.0 &20.3 &35.8 &21.0
    \\ 

& CReST~\cite{wei2021crest}     &34.14    &49.27   &24.06   &57.9 &51.5 &49.1 &22.7 &13.2 &66.2 &34.4 &39.4 &40.4 &24.6 &17.2 &10.2 &24.4 &36.5 &24.4
  \\

& SimiS~\cite{simis}   &36.89   &53.97   &25.46   &57.8 &58.6 &58.6 &22.9 &0.0 &70.9 &38.0 &52.0 &47.0 &32.4 &20.2 &11.5 &18.1 &39.9 &25.0
  \\

& Basak \textit{et al.}~\cite{basak2022addressing}    &29.84     &44.75    &19.90  &50.7 &47.7 &44.1 &21.1 &0.0 &61.8 &27.7 &38.1 &40.4 &21.8 &9.6 &9.5 &14.6 &36.5 &24.0
    \\
                                 
& CLD~\cite{lin2022cld}    &36.18     &53.50  &24.63  &55.8 &55.8 &59.1 &23.9 &0.0 &69.9 &38.2 &50.1 &44.5 &32.3 &18.9 &9.2 &18.8 &42.2 &24.0
      \\

& DHC~\cite{wang2023dhc} &38.23    &54.47   &27.41  &62.1 &59.5 &57.8 &25.0 &20.5 &66.0 &38.2 &51.3 &47.9 &26.8 &26.4 &7.0 &17.8 &43.2 &24.0
  \\


&A\&D~\cite{wang2024towards}  &32.87       &51.35   &20.56    &68.5 &56.2 &62.3 &18.5 &0.0 &62.9 &40.1 &51.1 &41.1 &32.2 &0.0 &0.0 &24.7 &18.1 &17.4\\

&GA~\cite{wu98gradient} &53.84     &66.70   &45.28   &72.1  &68.0 &72.4 &44.6 &42.7 &\textbf{82.7} &48.1 &66.3 &61.3 &49.5 &44.9 &30.4 &31.6 &56.9 &36.2 \\

 \midrule
 
& \cellcolor{cyan!20}TAK (Ours) &\cellcolor{cyan!20}\textbf{60.84}    &\cellcolor{cyan!20}\textbf{70.81}  &\cellcolor{cyan!20}\textbf{54.20 } &\cellcolor{cyan!20}\textbf{73.5} & \cellcolor{cyan!20}\textbf{72.9}
& \cellcolor{cyan!20}\textbf{74.7} & \cellcolor{cyan!20}\textbf{50.9}
& \cellcolor{cyan!20}\textbf{50.2} & \cellcolor{cyan!20}82.1
& \cellcolor{cyan!20}\textbf{58.2} & \cellcolor{cyan!20}\textbf{74.3}
& \cellcolor{cyan!20}\textbf{67.7} & \cellcolor{cyan!20}\textbf{57.4}
& \cellcolor{cyan!20}\textbf{49.2} & \cellcolor{cyan!20}\textbf{48.9}
& \cellcolor{cyan!20}\textbf{44.7} & \cellcolor{cyan!20}\textbf{63.5}
& \cellcolor{cyan!20}44.5   \\

\bottomrule

\end{tabular}
}
\caption{Quantitative comparison between our approach and SOTA SSL segmentation methods on {2\% labeled AMOS dataset~\cite{amos}}. `General' or `Imbalance' indicate whether the methods consider class-imbalance issue or not. 
}
\label{tab:amos2_baselin1}
\end{table*}
\begin{table*}[t]
\scriptsize
\resizebox*{\linewidth}{!}{

\begin{tabular}{c|c|ccc|ccccccccccccccc}
\toprule
\multicolumn{2}{c|}{\multirow{2}{*}{Methods}} & \multicolumn{3}{c|}{Avg.Dice}  & \multicolumn{15}{c}{Average Dice of Each   Class}                                        \\ 
\multicolumn{2}{c|}{}                          & All            &\cellcolor{orange!20}L.    &\cellcolor{green!20}S.     & \cellcolor{orange!20}Sp   & \cellcolor{orange!20}RK   &\cellcolor{orange!20} LK   &\cellcolor{green!20} Ga   &\cellcolor{green!20} Es   &\cellcolor{orange!20} Li   &\cellcolor{orange!20} St   &\cellcolor{green!20} Ao   &\cellcolor{green!20} IVC   &\cellcolor{green!20} PA   &\cellcolor{green!20} RAG  & \cellcolor{green!20}LAG  &\cellcolor{green!20} Du   &\cellcolor{orange!20} Bl  &\cellcolor{green!20} P/U  \\\midrule
\multirow{9}{*}{\rotatebox{90}{General}}         & V-Net (fully)  &76.50  
  &86.63    &69.74     &92.2 &92.2 &93.3 &65.5 &70.3 &95.3 &82.4 &91.4 &85.0 &74.9 &58.6 &58.1 &65.6 &64.4 &58.3   \\ 
\midrule

& UA-MT~\cite{yu2019uamt}  &42.16      &55.65    &33.18    &59.8 &64.9 &64.0 &35.3 &34.1 &77.7 &37.8 &61.0 &46.0 &33.3 &26.9 &12.3 &18.1 &29.7 &31.6     \\



& CPS~\cite{chen2021semi}   &41.08      &53.78    &32.61   &56.1 &60.3 &59.4 &33.3 &25.4 &73.8 &32.4 &65.7 &52.1 &31.1 &25.5 &6.2 &18.4 &40.7 &35.8 \\

&DeSCO~\cite{cai2023orthogonal} &44.40      &73.57    &24.97   &78.9 &81.4 &81.8 &6.7 &0.0 &88.2 &44.2 &78.9 &61.5 &37.2 &0.0 &0.0 &21.2 &66.9 &19.2 \\

& DePL~\cite{wang2022depl}    &41.98     &53.67    &34.19   &55.7 &62.4 &57.7 &36.6 &31.3 &68.4 &33.9 &65.6 &51.9 &30.2 &23.3 &10.2 &20.9 &43.9 &37.7    \\
&Co-BioNet~\cite{peiris2023uncertainty} &48.32     &71.05    &33.16    &76.6 &82.1 &75.1 &41.5 &38.2 &87.9 &40.4 &75.2 &53.7 &40.8 &4.8 &0.0 &25.1 &64.2 &19.2\\

&MagicNet~\cite{chen2023magicnet} &54.09    &74.88    &40.23    &80.0 &84.5 &86.1 &47.9 &0.0 &85.1 &50.7 &81.7 &69.3 &57.2 &46.0 &0.0 &40.8 &62.9 &19.2 \\

\midrule
\multirow{8}{*}{\rotatebox{90}{Imbalance}} 

& Adsh~\cite{guo2022adsh}  &40.32     &54.17    &31.09    &56.0 &63.6 &57.3 &34.7 &25.7 &73.9 &30.7 &65.7 &51.9 &27.1 &20.2 &0.0 &18.6 &43.5 &35.9       \\ 

& CReST~\cite{wei2021crest}  &46.56    &60.23    &37.44    &66.5 &64.2 &65.4 &36.0 &32.2 &77.8 &43.6 &68.5 &52.9 &40.3 &24.7 &19.5 &26.5 &43.9 &36.4     \\

& SimiS~\cite{simis}  &47.27     &66.57    &34.40   &77.4 &72.5 &68.7 &32.1 &14.7 &86.6 &46.3 &74.6 &54.2 &41.6 &24.4 &17.9 &21.9 &47.9 &28.2   \\

& Basak \textit{et al.}~\cite{basak2022addressing}    &38.74    &58.45    &25.60    &68.8 &59.0 &54.2 &29.0 &0.0 &83.7 &39.3 &61.7 &52.1 &34.6 &0.0 &0.0 &26.8 &45.7 &26.2    \\
                                 
& CLD~\cite{lin2022cld}   &46.10    &60.61    &36.42    &67.2 &68.5 &71.4 &41.0 &21.0 &76.1 &42.4 &69.8 &52.1 &37.9 &24.7 &23.4 &22.7 &38.1 &35.2       \\

& DHC~\cite{wang2023dhc}  &49.53     &61.80    &41.36    &68.1 &69.6 &71.1 &42.3 &37.0 &76.8 &43.8 &70.8 &57.4 &43.2 &27.0 &28.7 &29.1 &41.4 &36.7 \\


&A\&D~\cite{wang2024towards}  &37.83    &61.38   &22.13    &72.8 &67.5 &64.4 &14.6 &0.0 &82.3 &44.6 &70.7 &51.9 &38.1 &0.0 &0.0 &23.7 &36.7 &0.2
\\

&GA~\cite{wu98gradient} &63.50     &77.71    &54.03    &78.9 &85.5 &87.2 &50.0 &49.1 &86.9 &56.2 &83.4 &70.3 &57.4 &49.1 &40.8 &38.3 &71.6 &47.9 \\

 \midrule
& \cellcolor{cyan!20}TAK (Ours)  & \cellcolor{cyan!20}\textbf{70.20 }     &\cellcolor{cyan!20}\textbf{83.01}    &\cellcolor{cyan!20}\textbf{61.65}     &\cellcolor{cyan!20}\textbf{87.6} &\cellcolor{cyan!20}\textbf{88.4} &\cellcolor{cyan!20}\textbf{90.9} &\cellcolor{cyan!20}\textbf{57.8} &\cellcolor{cyan!20}\textbf{64.0} &\cellcolor{cyan!20}\textbf{91.4} &\cellcolor{cyan!20}\textbf{66.9} &\cellcolor{cyan!20}\textbf{87.7} &\cellcolor{cyan!20}\textbf{77.1} &\cellcolor{cyan!20}\textbf{64.9} &\cellcolor{cyan!20}\textbf{55.9} &\cellcolor{cyan!20}\textbf{49.3} &\cellcolor{cyan!20}\textbf{46.5} &\cellcolor{cyan!20}\textbf{72.9} &\cellcolor{cyan!20}\textbf{51.7}
\\


\bottomrule

\end{tabular}
}
\caption{Quantitative comparison between our approach and SOTA SSL segmentation methods on {5\% labeled AMOS dataset~\cite{amos}}. `General' or `Imbalance' indicate whether the methods consider class-imbalance issue or not. 
}
\label{tab:amos5_baselin1}
\end{table*}
For the visual branch, we use the mean-teacher framework to extract image features, where the teacher model weights are updated as an exponential moving average (EMA) of the student weights. 
We perform multi-scale contrastive learning on both text embeddings and visual embeddings encoded by the student vision network in the Cross-Modal Contrastive
Alignment Module. 
Let the multi-scale image features extracted by the student vision encoder be denoted as $ \{E^{i} \in \mathbb{R}^{C^{i} \times D^{i} \times W^{i} \times H^{i}}\}^{i \in (1,2,\ldots)}$, where $i$ represents different stages of the vision encoder. Here, \( C^{i} \) denotes the number of channels, \( D^{i} \) represents the depth, \( W^{i} \) is the width, and \( H^{i} \) is the height of the feature map at the \( i \)-th stage. For labeled data, we downsample the label to the appropriate size to extract features for each category. For unlabeled data, we estimate uncertainty of voxels using entropy, as described by the following formula:
\begin{equation}\label{entropy}
\mathcal{H}(\hat{p}_{ij}) = -\sum_{c=1}^{C} \hat{p}_{ij}^c \log \hat{p}_{ij}^c.
\end{equation}
We treat samples with higher entropy as uncertain samples and exclude them when selecting visual samples for contrastive learning. The set of visual features for the $k$-th class selected from both labeled and unlabeled data is denoted as $V_{k}$. By applying a multi-layer perceptron (MLP) to $T_{k}^{p}$, $T_{k}^{s}$, we map them into the corresponding dimensions, resulting in $\{T_{k}^{p,i}\in\mathbb{R}^{C^{i}}  \}^{i\in (1,2,\ldots)}$ and $\{T_{k}^{s,i} \in \mathbb{R}^{C^{i}}  \} ^{i\in (1,2,\ldots)}$. 
For $k$-th class, let $F_{k}^{i} = V_{k}^{i} \cup T_{k}^{p,i} \cup T_{k}^{s,i}$. 
For two embeddings $f_1 \in F_{k_{1}}^{i_1} $ and $f_2 \in F_{k_{2}}^{i_2} $, if $k_1=k_2$ and $i_1=i_2$,  then $f_1$ and $f_2$ form a positive pair; if $k_1 \neq k_2$ and $i_1 = i_2$,  then $f_1$ and $f_2$ form a negative pair. The local contrastive loss is defined as
\begin{equation}\label{loss_contrast}
\mathcal{L}_{con} = -\frac{1}{|\Omega|}  \sum_{f \in \Omega}log \frac{\sum\nolimits_{f_{p} \in P(f)} exp(f \cdot f_{p} / \tau)}{\sum\nolimits_{f_{n} \in N(f)} exp(f \cdot f_{n}/ \tau)}
\end{equation}
where $|\Omega|$ is the union of the visual embeddings and text embeddings. $P(f)$ and $N(f)$  denote the positive set and negative set of $f$.
\subsection{Training objective}
For each labeled image $x^l$, we adopt the cross-entropy loss $\ell_{ce}$ and dice loss $\ell_{dc}$ as the supervised loss $\mathcal{L}_s$ given by:
\begin{equation}\label{labeled supervise}
\mathcal{L}_s =  \ell_{ce}({p}^l, {y}^l) + \ell_{dc}({p}^l, {y}^l)
\end{equation}
where ${p}^l$ is the prediction output of the student networks, and $y^l$ is the corresponding label. For each unlabeled image $x^u$, we use the pseudo label obtained from the teacher network to supervise the output of another one. The loss $\mathcal{L}_u$ for the unlabeled image $x^u$ is given by:
\begin{equation}\label{unlabeled supervise}
\mathcal{L}_u = \ell_{ce}({p}^u, \hat{y}) + \ell_{dc}({p}^u, \hat{y}) 
\end{equation}
where $\hat{y}$ is the pseudo labels. The overall training objective $\mathcal{L}$ is defined by:
\begin{equation} \label{eq:overallloss}
     \mathcal{L} =\mathcal{L}_s + \lambda_{u} \times \mathcal{L}_u + \lambda_{c} \times \mathcal{L}_{con}, 
\end{equation}
 where $\lambda_{u}$ and $\lambda_{c}$ are the coefficients of $\mathcal{L}_u$ and $\mathcal{L}_{con}$. It is worth noting that the textual anatomical knowledge $T_{k}^{p}$ and $T_{k}^{s}$ only needs to be generated once before the entire training process begins.

\begin{table*}[t]
\scriptsize
\centering
\resizebox*{1\linewidth}{!}{
\begin{tabular}{c|c|ccc|ccccccccccccc}
\toprule
\multicolumn{2}{c|}{\multirow{2}{*}{Methods}}  & \multicolumn{3}{c|}{Avg.Dice} & \multicolumn{13}{c}{Average Dice of Each  Class}                                        \\ 
\multicolumn{2}{c|}{}       &All            &\cellcolor{orange!20}L.      &\cellcolor{green!20}S.     &\cellcolor{orange!20}Sp   &\cellcolor{orange!20}RK   &\cellcolor{orange!20}LK   &\cellcolor{green!20}Ga   &\cellcolor{green!20}Es   &\cellcolor{orange!20}Li   &\cellcolor{orange!20}St   &\cellcolor{green!20}Ao   &\cellcolor{green!20}IVC   &\cellcolor{green!20}PSV  &\cellcolor{green!20}PA   &\cellcolor{green!20}RAG   &\cellcolor{green!20}LAG   \\\midrule
\multirow{9}{*}{\rotatebox{90}{General}}         & V-Net (fully)   &68.49 & 88.64   &55.90 &90.2 &91.9 &90.7 &38.3 &30.9 &94.8 &75.6 &79.1 &81.4 &62.1 &48.5 &48.9 &58.0     \\ \midrule

& UA-MT~\cite{yu2019uamt}   &28.80 &46.80   &17.56  &37.0 &49.6 &29.1 &6.0 &11.5 &85.2 &33.1 &61.4 &34.2 &12.8 &5.5 &2.5 &6.6
 \\



& CPS~\cite{chen2021semi}  &30.28 &53.68   &15.66   &63.6 &46.1 &45.5 &0.0 &0.0 &74.5 &38.7 &64.3 &51.0 &0.0 &10.0 &0.0 &0.0 \\

& DeSCO~\cite{cai2023orthogonal}   &38.91  &73.26    &17.45  &68.7 &79.5 &76.5 &0.0 &0.0 &90.4 &51.2 &71.2 &59.3 &0.0 &9.1 &0.0 &0.0 \\

& DePL~\cite{wang2022depl}   &36.18  &53.94   &25.08   &54.9 &52.2 &48.3 &0.0 &30.2 &70.3 &44.0 &65.8 &46.2 &13.8 &13.5 &9.9 &21.3 \\

&Co-BioNet~\cite{peiris2023uncertainty} &40.84  & 62.62     &27.23    &59.5 &68.6 &52.5 &6.0 &30.0 &\textbf{91.1} &41.4 &72.0 &48.6 &13.6 &9.0 &10.6 &28.1\\

&MagicNet~\cite{chen2023magicnet} &54.38  &78.64    &39.22      &73.0 &83.8 &82.3 &13.2 &0.0 &90.9 &63.2 &78.3 &69.4 &47.1 &35.4 &23.7 &46.7  \\

\midrule
\multirow{8}{*}{\rotatebox{90}{Imbalance}} 

& Adsh~\cite{guo2022adsh}  &30.95  &54.76     &16.07     &62.0 &45.5 &40.6 &0.0 &0.0 &81.5 &44.2 &66.4 &45.5 &9.6 &7.1 &0.0 &0.0   \\ 

& CReST~\cite{wei2021crest}  &33.65  &48.50    &24.37   &47.2 &49.2 &39.1 &2.3 &23.2 &70.4 &36.6 &63.3 &37.9 &17.5 &13.3 &13.9 &23.6   \\

& SimiS~\cite{simis} &30.48 & 51.96   &17.05    &43.8 &62.1 &47.3 &5.6 &0.0 &59.0 &47.6 &68.8 &41.0 &8.6 &12.4 &0.0 &0.0\\

& Basak \textit{et al.}~\cite{basak2022addressing}  &34.40 &61.96  &17.17    &67.5 &62.3 &55.9 &0.0 &0.0 &81.4 &42.7 &66.9 &51.2 &12.1 &7.2 &0.0 &0.0  \\
                                 
& CLD~\cite{lin2022cld}  &35.12  &52.24  &24.41    &54.1 &55.2 &41.0 &12.2 &0.0 &67.5 &43.4 &71.6 &50.0 &18.1 &11.9 &3.3 &28.2 \\

& DHC~\cite{wang2023dhc} &36.92   & 53.08    &26.82   &57.0 &46.4 &39.9 &5.6 &20.6 &73.3 &48.8 &70.8 &50.6 &16.9 &10.3 &11.0 &28.8 \\


&A\&D~\cite{wang2024towards}  &50.11   &71.82     &36.55    &76.7 &71.9 &71.8 &4.2 &34.7 &86.6 &52.1 &68.1 &71.0 &32.8 &27.7 &22.8 &31.1 \\

&GA~\cite{wu98gradient}  &57.45   &78.00     &44.61     &69.8 &85.8 &83.1 &10.2 &\textbf{49.9} &90.6 &60.7 &76.4 &69.2 &41.8 &32.0 &29.3 &48.1\\

 \midrule
& \cellcolor{cyan!20}  TAK (Ours)  & \cellcolor{cyan!20}\textbf{65.75}   &\cellcolor{cyan!20}\textbf{81.32}  &\cellcolor{cyan!20}\textbf{56.03}     &\cellcolor{cyan!20}\textbf{78.5} &\cellcolor{cyan!20}\textbf{88.0} &\cellcolor{cyan!20}\textbf{85.6} &\cellcolor{cyan!20}\textbf{26.2} &\cellcolor{cyan!20}\textbf{49.9} &\cellcolor{cyan!20}89.7 &\cellcolor{cyan!20}\textbf{64.8} &\cellcolor{cyan!20}\textbf{79.2} &\cellcolor{cyan!20}\textbf{79.4} &\cellcolor{cyan!20}\textbf{58.6} &\cellcolor{cyan!20}\textbf{46.5} &\cellcolor{cyan!20}\textbf{49.1} &\cellcolor{cyan!20}\textbf{59.3}
\\

\bottomrule

\end{tabular}
}
\caption{Quantitative comparison between our approach and SOTA SSL segmentation methods on {10\% labeled Synapse dataset~\cite{synapse}}. `General' or `Imbalance' indicate whether the methods consider class-imbalance issue or not. 
}
\label{tab:synaspe10_baseline1}

\end{table*}
\begin{table*}[t]
\scriptsize
\centering
\resizebox*{1\linewidth}{!}{
\begin{tabular}{c|c|ccc|ccccccccccccc}
\toprule
\multicolumn{2}{c|}{\multirow{2}{*}{Methods}}  & \multicolumn{3}{c|}{Avg.Dice}& \multicolumn{13}{c}{Average Dice of Each   Class}                                        \\ 
\multicolumn{2}{c|}{}                          & ALL            &\cellcolor{orange!20}L.         &\cellcolor{green!20}S.               &\cellcolor{orange!20}Sp   &\cellcolor{orange!20}RK   &\cellcolor{orange!20}LK   &\cellcolor{green!20}Ga   &\cellcolor{green!20}Es   &\cellcolor{orange!20}Li   &\cellcolor{orange!20}St   &\cellcolor{green!20}Ao   &\cellcolor{green!20}IVC   &\cellcolor{green!20}PSV  &\cellcolor{green!20}PA   &\cellcolor{green!20}RAG   &\cellcolor{green!20}LAG   \\\midrule
\multirow{9}{*}{\rotatebox{90}{General}}         & V-Net (fully)   &68.49  &88.64    &55.90    &90.2 &91.9 &90.7 &38.3 &30.9 &94.8 &75.6 &79.1 &81.4 &62.1 &48.5 &48.9 &58.0 \\
\midrule

& UA-MT~\cite{yu2019uamt}  &41.37 & 69.56   &23.75    &75.2 &81.0 &66.8 &0.0 &0.0 &86.9 &37.9 &69.4 &67.8 &31.1 &21.7 &0.0 &0.0
     \\



& CPS~\cite{chen2021semi}   &48.50 &80.00  &28.81   &83.9 &87.8 &85.8 &0.0 &0.0 &92.3 &50.2 &75.0 &74.3 &55.9 &25.3 &0.0 &0.0 \\
&DeSCO~\cite{cai2023orthogonal}  &44.46  &79.56    &22.52  &82.4 &89.4 &87.4 &0.0 &0.0 &89.0 &49.6 &75.3 &76.3 &1.8 &26.8 &0.0 &0.0 \\ 

& DePL~\cite{wang2022depl}   &59.44  &81.42  &45.71   &84.4 &87.4 &85.7 &5.5 &22.1 &90.9 &58.7 &75.4 &77.4 &55.8 &37.4 &43.5 &48.6     \\
&Co-BioNet~\cite{peiris2023uncertainty} &58.83  &79.86  &45.70  &82.8 &90.0 &86.5 &11.6 &19.5 &\textbf{92.3} &47.7 &77.5 &77.7 &51.3 &30.3 &47.5 &50.2\\

&MagicNet~\cite{chen2023magicnet} &60.57  &83.02  &46.54   &82.5 &91.0 &89.5 &11.2 &0.0 &89.4 &62.7 &77.6 &79.0 &66.1 &47.3 &36.8 &54.3\\

\midrule
\multirow{8}{*}{\rotatebox{90}{Imbalance}} 

& Adsh~\cite{guo2022adsh}      &44.06  &72.94 &26.01   &77.2 &81.2 &77.1 &0.0 &0.0 &86.1 &43.1 &70.7 &71.8 &43.7 &21.9 &0.0 &0.0
   \\ 

& CReST~\cite{wei2021crest}   &59.99  &78.00   &48.73   &77.3 &87.6 &85.6 &19.4 &36.5 &90.0 &49.5 &76.3 &72.6 &51.0 &37.6 &43.3 &53.2    \\

& SimiS~\cite{simis}  &50.46  &80.10    &31.93   &83.3 &90.8 &85.8 &9.2 &0.0 &85.6 &55.0 &73.6 &71.7 &50.4 &34.0 &0.0 &16.6  \\

& Basak \textit{et al.}~\cite{basak2022addressing}  &48.38  &79.22   & 29.11   &84.6 &86.9 &79.8 &0.0 &0.0 &90.2 &54.6 &72.6 &73.2 &55.5 &31.6 &0.0 &0.0      \\
                                 
& CLD~\cite{lin2022cld}    &49.47  &78.14    &31.55   &83.3 &86.7 &85.7 &1.3 &0.0 &85.9 &49.1 &74.5 &76.3 &52.4 &33.8 &14.1 &0.0      \\
                                 
& DHC~\cite{wang2023dhc} &58.97  &79.02  &46.44   &81.6 &87.5 &85.5 &12.4 &27.4 &88.8 &51.7 &74.3 &73.7 &55.2 &33.3 &46.1 &49.1   \\


&A\&D~\cite{wang2024towards} &60.88  & 72.16   &53.83   &\textbf{85.2} &66.9 &67.0 &\textbf{52.7} &62.9 &89.6 &52.1 &\textbf{83.0 } &74.9 &41.8 &43.4 &44.8 &27.2 \\

&GA~\cite{wu98gradient} &68.43  &82.92    &59.38   &81.4 &92.4 &\textbf{90.8} &33.5 &53.3 &89.1 &60.9 &79.1 &82.1 &66.7 &48.7 &50.3 &61.4
 \\

 \midrule
 & \cellcolor{cyan!20}TAK (Ours) & \cellcolor{cyan!20}\textbf{72.66}    & \cellcolor{cyan!20}\textbf{85.78}  & \cellcolor{cyan!20}\textbf{64.46}  & \cellcolor{cyan!20}84.0 & \cellcolor{cyan!20}\textbf{92.9} & \cellcolor{cyan!20}87.4 & \cellcolor{cyan!20}44.0 & \cellcolor{cyan!20}\textbf{57.4} & \cellcolor{cyan!20}91.0 & \cellcolor{cyan!20}\textbf{73.6} & \cellcolor{cyan!20}81.0 & \cellcolor{cyan!20}\textbf{83.1} & \cellcolor{cyan!20}\textbf{70.6} & \cellcolor{cyan!20}\textbf{57.0} & \cellcolor{cyan!20}\textbf{57.5} & \cellcolor{cyan!20}\textbf{65.1}
 \\

\bottomrule

\end{tabular}
}

\caption{Quantitative comparison between our approach and SOTA SSL segmentation methods on {20\% labeled Synapse dataset~\cite{synapse}}. `General' or `Imbalance' indicate whether the methods consider class-imbalance issue or not. 
}
\label{tab:synapse20_baseline1}
\end{table*}

\begin{table*}[t]
  \centering
  \setlength{\tabcolsep}{1mm}
    \resizebox{1\linewidth}{!}{
  \begin{tabular}{ccc|c|ccc|ccccccccccccccc}
    \hline
    \multicolumn{3}{c|}{Prompt}      &\multirow{2}{*}{Contrast}   &\multicolumn{3}{c|}{Avg.}   & \multicolumn{15}{c}{Dice of Each  Class}  \\
    Name  &Positon &Shape  &  &All &\cellcolor{orange!20}L. & \cellcolor{green!20}S.   &\cellcolor{orange!20} Sp   
    &\cellcolor{orange!20}RK   & \cellcolor{orange!20}LK   &\cellcolor{green!20} Ga   &\cellcolor{green!20} Es   &\cellcolor{orange!20} Li   & \cellcolor{orange!20}St   &\cellcolor{green!20} Ao   & \cellcolor{green!20}IVC   & \cellcolor{green!20}PA   &\cellcolor{green!20} RAG  &\cellcolor{green!20} LAG  & \cellcolor{green!20}Du   & \cellcolor{orange!20} Bl  &\cellcolor{green!20} P/U   \\
    \hline
    \checkmark  & &  &  &66.04 &79.43  &57.12   &82.4 &86.7 &89.0 &51.8 &59.9 &88.3 &60.8 &84.5 &70.6 &59.1 &50.4 &47.6 &39.6 &69.4 &50.6  
    \\
   &\checkmark  &   & & 67.49 &80.93 &58.53  &85.2 &87.8 &89.6 &52.9 &59.7 &89.5 &61.0 &85.8 &73.2  &61.7 &50.8 &49.1 &42.7 &72.5 &50.9  \\
      &  &\checkmark   &  &67.36 &80.93  & 58.31   &83.2 &87.6 &89.8 &54.6 &59.1 &88.7 &61.7 &85.2 &71.0 &62.1 &50.5 &49.1 &41.2 &74.6 &52.0 \\

    &\checkmark  &\checkmark  &  &68.16 &80.68  &59.81   &82.7 &87.4 &89.8 &51.9 &61.5 &88.7 &64.6 &87.1 &75.4 &63.6 &54.3 &50.9 &41.5 &70.9 &52.1  \\

    \checkmark & &  &\checkmark    &67.26  &81.86 &57.52  & 85.7 & 88.4 & 89.9 & 52.3 & 59.8 & 90.1 & 60.0 & 84.7 & 73.3 & 57.7 & 51.9 & 47.2 & 39.7 & 77.1 & 51.1 \\

      &\checkmark &\checkmark  &\checkmark   &70.20 &83.01 &61.65    &87.6 &88.4 &90.9 &57.8 &64.0 &91.4 &66.9 &87.7 &77.1 &64.9 &55.9 &49.3 &46.5 &72.9 &51.7  \\

    \hline
  \end{tabular}
  }
    \caption{Ablation study results evaluating various textual organ descriptors as prompts and the incorporation of the text and visual alignment module using 5\% labeled AMOS dataset.}
  \label{tab:abaltion}
\end{table*}

\section{Experiments}
\subsection{Dataset and evaluation protocal}
 \noindent \textbf{AMOS.} The AMOS dataset~\cite{amos} consists of 360 scans.
 Following the experimental setup proposed in DHC~\cite{wang2023dhc}, we divide 360 scans into 216, 24 and 120 scans for training, validation, and testing.
 \textbf{Synapse.} The Synapse dataset~\cite{synapse} consists of 30 scans. 
 Following data setting in~\cite{wang2023dhc}, we split 30 scans as 20, 4 and 6 scans for training, validation, and testing, respectively. The proposed method is evaluated with two widely used metrics in semi-supervised medical image segmentation: Dice coefficient (Dice) and the average surface distance (ASD).

\subsection{Implementation details}
We conducted all experiments on a single NVIDIA A100 GPU (40G). we use MagicNet~\cite{chen2023magicnet} as the backbone of the vision branch. The proposed method named TAK is trained using the SGD  optimizer with an initial learning rate of 0.01, momentum of 0.9 and decay
of $10e^{-4}$. Following GA~\cite{wu98gradient}, we randomly cropped patches of size $96\times96\times96$ during training. No additional data augmentation is applied beyond random cropping. The batch size is set to 4, with 2 labeled patches and 2 unlabeled patches. During the final testing phase, a sliding window approach is used, employing a stride of $32\times32\times16$. 
The contrastive learning loss is applied starting from the 20th epoch, with a contrastive learning loss coefficient $\lambda_{c} =  0.1$. We adopt the same setting for $\lambda_{u}$ in GA~\cite{wu98gradient}.

\begin{figure}[tb]
    \centering
    \begin{subfigure}[b]{0.2\linewidth}
        \centering
        \includegraphics[width=\linewidth]{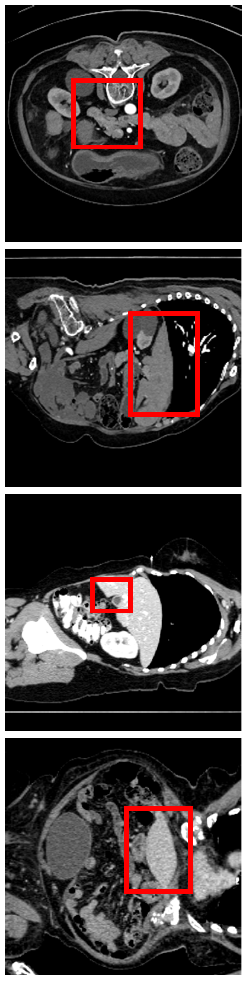}
        \caption{Image}
    \end{subfigure}\hspace{-1mm}
    \begin{subfigure}[b]{0.2\linewidth}
        \centering
        \includegraphics[width=\linewidth]{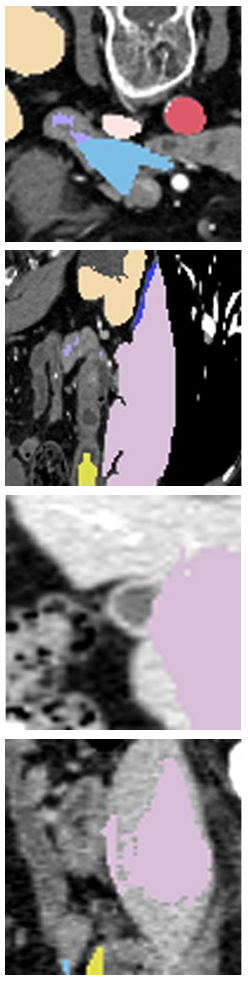}
        \caption{A\&D~\cite{wang2024towards}}
    \end{subfigure}\hspace{-1mm}
    \begin{subfigure}[b]{0.2\linewidth}
        \centering
        \includegraphics[width=\linewidth]{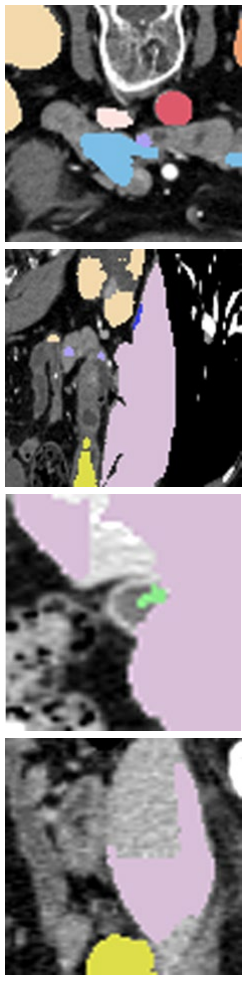}
        \caption{GA~\cite{wu98gradient}}
    \end{subfigure}\hspace{-1mm}
    \begin{subfigure}[b]{0.2\linewidth}
        \centering
        \includegraphics[width=\linewidth]{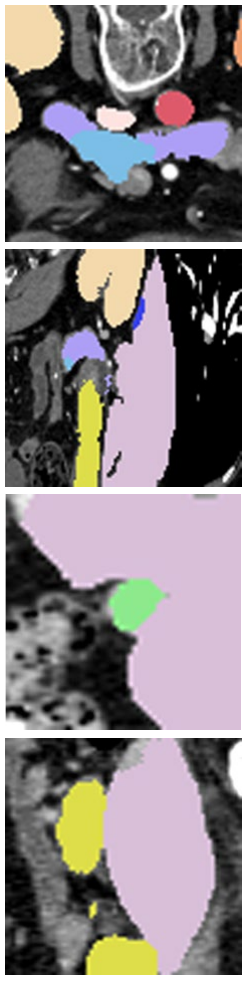}
        \caption{TAK}
    \end{subfigure}\hspace{-1mm}
    \begin{subfigure}[b]{0.2\linewidth}
        \centering
        \includegraphics[width=\linewidth]{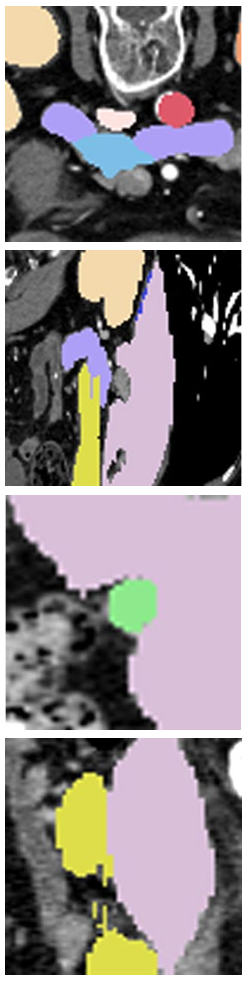}
        \caption{GT}
    \end{subfigure}\hspace{-1mm}
    
    \caption{Some qualitative segmentation results of our TAK and some state-of-the-art methods on Amos dataset~\cite{amos} (first two rows) and Synapse dataset~\cite{synapse} (last two rows).}
    \label{fig:view}
\end{figure}

\subsection{Comparative results on different datasets}
Following GA~\cite{wu98gradient}, we evaluate our approach against six state-of-the-art (SOTA) semi-supervised segmentation methods -- UA-MT~\cite{yu2019uamt}, CPS~\cite{chen2021semi}, DeSCO~\cite{cai2023orthogonal}, DePL~\cite{wang2022depl}, Co-BioNet~\cite{peiris2023uncertainty}, and MagicNet~\cite{chen2023magicnet} -- as well as eight advanced SOTA techniques for handling class imbalance: Adsh~\cite{guo2022adsh}, CReST~\cite{wei2021crest}, SimiS~\cite{simis}, Basak et al.~\cite{basak2022addressing}, CLD~\cite{lin2022cld}, DHC~\cite{wang2023dhc}, A\&D~\cite{wang2024towards}, and GA~\cite{wu98gradient}. 
We define classes with a voxel proportion of less than 5\% of all organ categories (excluding the background) as small classes and those exceeding 5\% as large classes. The specific voxel proportions of different organs can be found in the supplementary. In AMOS dataset~\cite{amos}, large classes include liver (Li), stomach (St), spleen (Sp), left kidney (LK), right kidney (RK) , and bladder (Bl); small classes include the rest. In Synapse dataset~\cite{synapse}, large classes include Li, St, Sp, LK and RK; the others are small classes. 
Compared to various state-of-the-art methods, our proposed TAK achieves significant improvements. Some qualitative results are shown in Fig.~\ref{fig:view}. Fig.~\ref{tab:run_time} compares the computational complexity and model size of different methods, showing TAK's significant improvements with similar FLOPs and parameters as GA~\cite{wu98gradient}.
The results using the ASD metric and is provided in the supplementary.

 \begin{table}[t]
      \centering
\resizebox{0.8\linewidth}{!}{
     \begin{tabular}{c c c c c}
     \toprule
        Method  &DHC~\cite{wang2023dhc} &Co-BioNet~\cite{peiris2023uncertainty} &GA~\cite{wu98gradient}  &TAK\\
    \midrule
      FLOPs (G)    &566.76 &285.68  &143.40  &150.96    \\
      Params (M)    &37.8    &18.8   &18.0  &18.5   \\
    \bottomrule
     \end{tabular}
     }
     \caption{Comparison of computational complexity (FLOPs) and model size (Params) among some different state-of-the-art method and our TAK. }
     \label{tab:run_time}
 \end{table}

\noindent \textbf{Comparative results on AMOS dataset}: As shown in Table~\ref{tab:amos2_baselin1}, in the scenario with 2\% labeled data from the AMOS dataset~\cite{amos}, TAK achieves a 22.61\% improvement in the Dice coefficient compared to DHC~\cite{wang2023dhc}. When compared to GA~\cite{wu98gradient}, the Dice coefficient increases by 7.00\%. Specifically, for large organs, the mean Dice score improves by 4.11\%, while for small organs, the improvement is 8.92\%.
As shown in Tab.~\ref{tab:amos5_baselin1}, with 5\% labeled data from the AMOS dataset~\cite{amos}, our TAK method enhances the Dice coefficient of GA~\cite{wu98gradient} from 63.50\% to 70.20\%. 
Specifically, our TAK improves GA~\cite{wu98gradient} by 5.30\% and  7.62\% for large and small organs, respectively.

\noindent \textbf{Comparative results on Synapse dataset}:
As shown in Tab.~\ref{tab:synaspe10_baseline1}, in the scenario with 10\% labeled data from the Synapse dataset~\cite{synapse}, TAK  achieves a 8.30\% improvement in average Dice coefficient compared to GA~\cite{wu98gradient}. Specifically, for large organs, the mean Dice score improves by 3.32\%, while for small organs, the improvement is 11.42\%. As shown in Tab.~\ref{tab:synapse20_baseline1}, in the scenario with 20\% labeled data, TAK enhances the Dice coefficient of GA~\cite{wu98gradient} from 68.43\% to 72.66\%. 

\subsection{Ablation studies}

\noindent \textbf{Ablation study on the effectiveness of using textual anatomical knowledge as prompts.}
To validate the effectiveness of using textual anatomical knowledge as prompts, we conduct ablation experiments with different descriptions as prompts. The experimental results are shown in Tab.~\ref{tab:abaltion}. `Name' represents using `A computerized tomography scan of the human abdomen includes the [CLS]', `Positions' refers to the description of the inter-organ relative positions priors , and `Shape' refers to the description of the organ shape prior. The Dice coefficient improves by 2.94\% (2.12\%) when both `Shape' and `Position' are used as descriptions, compared to using the `Name', with (or without) the Cross-Modal Contrastive
Alignment Module. This demonstrates the effectiveness of textual anatomical knowledge. It is worth noting that the descriptions denoted as `Shape' or `Positions' inherently encompass the information contained in the `Name' descriptor. We also conducted experiments combining all three descriptors together, which yielded experimental results that are nearly identical to those obtained when using only the `Shape' and `Position' descriptions.

\noindent \textbf{Ablation study on text and visual alignment module.}
As shown in Tab.~\ref{tab:abaltion}, `Contrast' means whether the Cross-Modal Contrastive
Alignment Module is used. When `Position' and `Shape' are used as text descriptions, the Cross-Modal Contrastive Alignment Module improves the Dice coefficient by 2.04\%.
Furthermore, We conduct an ablation study on the coefficient $\lambda_{c}$ of the contrastive learning loss, and the results are shown in the Tab.~\ref{table:con_w}. Additionally, we perform an ablation study to analyze the impact of the number of samples $\lambda_N$ drawn from the visual branch during contrastive loss computation, and the results are shown in the Tab.~\ref{table:con_sample}. We choose $\lambda_{c} = 0.1$ and a sample number $\lambda_N = 40$  as the hyperparameters.
 \begin{table}[t]
      \centering
\resizebox{1\linewidth}{!}{
     \begin{tabular}{c c c c c c c c c}
     \toprule
     $\lambda_{c}$  & 0 &0.01 &0.05 &0.1  &0.5 &1  &5 &10\\
    \midrule
        Dice (\%)  &68.16  &68.22  &69.02  &70.20  &70.18  &69.66  &68.78 &68.26\\
    \bottomrule
     \end{tabular}
     }
     \caption{Ablation study on the coefficient $\lambda_{c}$  of the contrastive learning loss using 5\% labeled AMOS dataset~\cite{amos}. }
     \label{table:con_w}
 \end{table}
 \begin{table}[t]
      \centering
\resizebox{1\linewidth}{!}{
     \begin{tabular}{c c c c c c c c c}
     \toprule
        $\lambda_N$  &0 &4 &12 &20 &40 &60 &80 \\
    \midrule
        Dice (\%)  &68.16  &69.05  &69.32  &69.42  &70.20  &69.48 &69.31   \\
    \bottomrule
     \end{tabular}
     }
     \caption{Ablation study on  the number of samples drawn from the visual branch during contrastive learning loss calculation using 5\% labeled AMOS dataset~\cite{amos}. }
     \label{table:con_sample}
 \end{table}

\noindent \textbf{Ablation study on extracting text embeddings using different pre-trained vision-language models.}
 We use three generalist models trained entirely on public medical datasets—UniMed-CLIP~\cite{khattak2024unimed}, PMC-CLIP~\cite{lin2023pmc}, and BioMedCLIP~\cite{zhang2023biomedclip}—along with the original CLIP~\cite{radford2021learning}. Results in Tab.~\ref{tab:different_clip} demonstrate that while there are slight performance variations among different CLIP variants, all models significantly outperform the baseline. Finally, we choose BioMedCLIP~\cite{zhang2023biomedclip} to extract textual embeddings of textual anatomical knowledge.
\begin{table}[t]
  \centering
  \setlength{\tabcolsep}{1mm}
   \resizebox{1\linewidth}{!}{
  \begin{tabular}{c  @{\hspace{1cm}}c  @{\hspace{1cm}}c @{\hspace{1cm}}c  @{\hspace{1cm}}c }

    \toprule
     &CLIP~\cite{radford2021learning}    &PMC-CLIP~\cite{lin2023pmc}  &UniMed-CLIP~\cite{khattak2024unimed}  &BioMedCLIP~\cite{zhang2023biomedclip} \\
    \midrule
        Dice   &69.23   &69.87   &69.53   &70.20  \\

    \bottomrule

   \end{tabular} 
    }
      \caption{Ablation study results using different pre-trained vision-language models to extract textual embeddings using 5\% labeled AMOS dataset~\cite{amos}.}
  \label{tab:different_clip}
\end{table}
\section{Conclusions}
In this work, we propose a novel semi-supervised framework for class-imbalanced multi-organ segmentation by integrating anatomical priors derived from multimodal large language models (MLLMs). By leveraging GPT-4o to generate structured textual descriptions of inter-organ spatial relationships and organ shape characteristics, our method transforms implicit anatomical knowledge into explicit, model-actionable priors. These priors are encoded as adaptive parameters in the segmentation head and aligned with visual features via cross-modal contrastive learning. Experiments demonstrate that our approach significantly improves segmentation accuracy for challenging small organs and morphologically complex structures, outperforming state-of-the-art methods. While our work focuses on anatomical spatial and shape priors, future research could extend this paradigm to incorporate other types of medical prior knowledge, such as pathological correlations, or physiological dynamics, to enhance segmentation tasks.
\clearpage
{
    \small
    \bibliographystyle{ieeenat_fullname}
    \bibliography{main}

\begin{thebibliography}{48}
\providecommand{\natexlab}[1]{#1}
\providecommand{\url}[1]{\texttt{#1}}
\expandafter\ifx\csname urlstyle\endcsname\relax
  \providecommand{\doi}[1]{doi: #1}\else
  \providecommand{\doi}{doi: \begingroup \urlstyle{rm}\Url}\fi

\bibitem[Alonso et~al.(2021)Alonso, Sabater, Ferstl, Montesano, and Murillo]{alonso2021semi}
Inigo Alonso, Alberto Sabater, David Ferstl, Luis Montesano, and Ana~C Murillo.
\newblock Semi-supervised semantic segmentation with pixel-level contrastive learning from a class-wise memory bank.
\newblock In \emph{Int. Conf. Comput. Vis.}, pages 8219--8228, 2021.

\bibitem[Basak and Yin(2023)]{basak2023pseudo}
Hritam Basak and Zhaozheng Yin.
\newblock Pseudo-label guided contrastive learning for semi-supervised medical image segmentation.
\newblock In \emph{IEEE Conf. Comput. Vis. Pattern Recog.}, pages 19786--19797, 2023.

\bibitem[Basak et~al.(2022)Basak, Ghosal, and Sarkar]{basak2022addressing}
Hritam Basak, Sagnik Ghosal, and Ram Sarkar.
\newblock Addressing class imbalance in semi-supervised image segmentation: A study on cardiac mri.
\newblock In \emph{Proc. of MICCAI}, pages 224--233, 2022.

\bibitem[Bloch et~al.(2003)Bloch, G{\'e}raud, and Ma{\^\i}tre]{bloch2003representation}
Isabelle Bloch, Thierry G{\'e}raud, and Henri Ma{\^\i}tre.
\newblock Representation and fusion of heterogeneous fuzzy information in the 3d space for model-based structural recognition—application to 3d brain imaging.
\newblock \emph{Artificial Intelligence}, 148\penalty0 (1-2):\penalty0 141--175, 2003.

\bibitem[Byrd and Lipton(2019)]{byrd2019effect}
Jonathon Byrd and Zachary Lipton.
\newblock What is the effect of importance weighting in deep learning?
\newblock In \emph{Proc. of Intl. Conf. on Machine Learning}, pages 872--881, 2019.

\bibitem[Cai et~al.(2023)Cai, Li, Qi, Yu, Shi, and Gao]{cai2023orthogonal}
Heng Cai, Shumeng Li, Lei Qi, Qian Yu, Yinghuan Shi, and Yang Gao.
\newblock Orthogonal annotation benefits barely-supervised medical image segmentation.
\newblock In \emph{IEEE Conf. Comput. Vis. Pattern Recog.}, pages 3302--3311, 2023.

\bibitem[Cao et~al.(2019)Cao, Wei, Gaidon, Arechiga, and Ma]{cao2019learning}
Kaidi Cao, Colin Wei, Adrien Gaidon, Nikos Arechiga, and Tengyu Ma.
\newblock Learning imbalanced datasets with label-distribution-aware margin loss.
\newblock \emph{Adv. Neural Inform. Process. Syst.}, 32, 2019.

\bibitem[Chen et~al.(2023)Chen, Bai, Shen, Li, Yu, and Wang]{chen2023magicnet}
Duowen Chen, Yunhao Bai, Wei Shen, Qingli Li, Lequan Yu, and Yan Wang.
\newblock Magicnet: Semi-supervised multi-organ segmentation via magic-cube partition and recovery.
\newblock In \emph{IEEE Conf. Comput. Vis. Pattern Recog.}, pages 23869--23878, 2023.

\bibitem[Chen et~al.(2022{\natexlab{a}})Chen, Fan, Wang, Wang, Schiele, Xie, Savvides, and Raj]{simis}
Hao Chen, Yue Fan, Yidong Wang, Jindong Wang, Bernt Schiele, Xing Xie, Marios Savvides, and Bhiksha Raj.
\newblock An embarrassingly simple baseline for imbalanced semi-supervised learning.
\newblock \emph{arXiv preprint arXiv:2211.11086}, 2022{\natexlab{a}}.

\bibitem[Chen et~al.(2021)Chen, Yuan, Zeng, and Wang]{chen2021semi}
Xiaokang Chen, Yuhui Yuan, Gang Zeng, and Jingdong Wang.
\newblock Semi-supervised semantic segmentation with cross pseudo supervision.
\newblock In \emph{IEEE Conf. Comput. Vis. Pattern Recog.}, pages 2613--2622, 2021.

\bibitem[Chen et~al.(2022{\natexlab{b}})Chen, Mancini, Zhu, and Akata]{chen2022semi}
Yanbei Chen, Massimiliano Mancini, Xiatian Zhu, and Zeynep Akata.
\newblock Semi-supervised and unsupervised deep visual learning: A survey.
\newblock \emph{IEEE Trans. on Pattern Anal. and Mach. Intell.}, 46\penalty0 (3):\penalty0 1327--1347, 2022{\natexlab{b}}.

\bibitem[Gu et~al.(2024)Gu, Liu, Sun, Zhu, Xu, and Najman]{gu2024shape}
Yuliang Gu, Yepeng Liu, Zhichao Sun, Jinchi Zhu, Yongchao Xu, and Laurent Najman.
\newblock Shape transformation driven by active contour for class-imbalanced semi-supervised medical image segmentation.
\newblock In \emph{2024 IEEE BIBM}, pages 1966--1973, 2024.

\bibitem[Gu et~al.(2025)Gu, Sun, Chen, Xiao, Liu, Xu, and Najman]{gu2025dual}
Yuliang Gu, Zhichao Sun, Tian Chen, Xin Xiao, Yepeng Liu, Yongchao Xu, and Laurent Najman.
\newblock Dual structure-aware image filterings for semi-supervised medical image segmentation.
\newblock \emph{Medical Image Analysis}, 99:\penalty0 103364, 2025.

\bibitem[Guo and Li(2022)]{guo2022adsh}
Lan-Zhe Guo and Yu-Feng Li.
\newblock Class-imbalanced semi-supervised learning with adaptive thresholding.
\newblock In \emph{Proc. of Intl. Conf. on Machine Learning}, pages 8082--8094, 2022.

\bibitem[Gupta et~al.(2022)Gupta, Hu, Kaan, Jin, Mpoy, Chung, Singh, Saltz, Kurc, Saltz, et~al.]{gupta2022learning}
Saumya Gupta, Xiaoling Hu, James Kaan, Michael Jin, Mutshipay Mpoy, Katherine Chung, Gagandeep Singh, Mary Saltz, Tahsin Kurc, Joel Saltz, et~al.
\newblock Learning topological interactions for multi-class medical image segmentation.
\newblock In \emph{Eur. Conf. Comput. Vis.}, pages 701--718, 2022.

\bibitem[Gupta et~al.(2024)Gupta, Zhang, Hu, Prasanna, and Chen]{gupta2024topology}
Saumya Gupta, Yikai Zhang, Xiaoling Hu, Prateek Prasanna, and Chao Chen.
\newblock Topology-aware uncertainty for image segmentation.
\newblock \emph{Adv. Neural Inform. Process. Syst.}, 36, 2024.

\bibitem[Huang et~al.(2024)Huang, Jiang, Zhang, Zhang, and Zhang]{huang2024cat}
Zhongzhen Huang, Yankai Jiang, Rongzhao Zhang, Shaoting Zhang, and Xiaofan Zhang.
\newblock Cat: Coordinating anatomical-textual prompts for multi-organ and tumor segmentation.
\newblock \emph{Adv. Neural Inform. Process. Syst.}, 2024.

\bibitem[Ji et~al.(2022)Ji, Bai, Yang, Ge, Zhu, Zhang, Li, Zhang, Ma, Wan, et~al.]{amos}
Yuanfeng Ji, Haotian Bai, Jie Yang, Chongjian Ge, Ye Zhu, Ruimao Zhang, Zhen Li, Lingyan Zhang, Wanling Ma, Xiang Wan, et~al.
\newblock Amos: A large-scale abdominal multi-organ benchmark for versatile medical image segmentation.
\newblock \emph{arXiv preprint arXiv:2206.08023}, 2022.

\bibitem[Jiang et~al.(2024)Jiang, Huang, Zhang, Zhang, and Zhang]{jiang2024zept}
Yankai Jiang, Zhongzhen Huang, Rongzhao Zhang, Xiaofan Zhang, and Shaoting Zhang.
\newblock Zept: Zero-shot pan-tumor segmentation via query-disentangling and self-prompting.
\newblock In \emph{IEEE Conf. Comput. Vis. Pattern Recog.}, pages 11386--11397, 2024.

\bibitem[Khattak et~al.(2024)Khattak, Kunhimon, Naseer, Khan, and Khan]{khattak2024unimed}
Muhammad~Uzair Khattak, Shahina Kunhimon, Muzammal Naseer, Salman Khan, and Fahad~Shahbaz Khan.
\newblock Unimed-clip: Towards a unified image-text pretraining paradigm for diverse medical imaging modalities.
\newblock \emph{arXiv preprint arXiv:2412.10372}, 2024.

\bibitem[Landman et~al.(2015)Landman, Xu, Igelsias, Styner, Langerak, and Klein]{synapse}
B Landman, Z Xu, J~Eugenio Igelsias, M Styner, T Langerak, and A Klein.
\newblock Miccai multi-atlas labeling beyond the cranial vault--workshop and challenge.
\newblock In \emph{Proc. MICCAI: Multi-Atlas Labeling Beyond Cranial Vault-Workshop Challenge}, 2015.

\bibitem[Li et~al.(2022)Li, Zhang, Zhang, Yang, Li, et~al.]{li2022grounded}
Liunian~Harold Li, Pengchuan Zhang, Haotian Zhang, Jianwei Yang, Chunyuan Li, et~al.
\newblock Grounded language-image pre-training.
\newblock In \emph{IEEE Conf. Comput. Vis. Pattern Recog.}, pages 10965--10975, 2022.

\bibitem[Li et~al.(2020)Li, Zhang, and He]{li2020shape}
Shuailin Li, Chuyu Zhang, and Xuming He.
\newblock Shape-aware semi-supervised 3d semantic segmentation for medical images.
\newblock In \emph{Proc. of MICCAI}, pages 552--561, 2020.

\bibitem[Lin et~al.(2023)Lin, Zhao, Zhang, Wu, Zhang, Wang, and Xie]{lin2023pmc}
Weixiong Lin, Ziheng Zhao, Xiaoman Zhang, Chaoyi Wu, Ya Zhang, Yanfeng Wang, and Weidi Xie.
\newblock Pmc-clip: Contrastive language-image pre-training using biomedical documents.
\newblock In \emph{International Conference on Medical Image Computing and Computer-Assisted Intervention}, pages 525--536, 2023.

\bibitem[Lin et~al.(2022)Lin, Yao, Li, Zheng, and Li]{lin2022cld}
Yiqun Lin, Huifeng Yao, Zezhong Li, Guoyan Zheng, and Xiaomeng Li.
\newblock Calibrating label distribution for class-imbalanced barely-supervised knee segmentation.
\newblock In \emph{Proc. of MICCAI}, pages 109--118, 2022.

\bibitem[Liu et~al.(2024)Liu, Li, Wu, and Lee]{liu2024visual}
Haotian Liu, Chunyuan Li, Qingyang Wu, and Yong~Jae Lee.
\newblock Visual instruction tuning.
\newblock \emph{Adv. Neural Inform. Process. Syst.}, 36, 2024.

\bibitem[Liu et~al.(2023)Liu, Zhang, Chen, Xiao, Lu, A~Landman, Yuan, Yuille, Tang, and Zhou]{liu2023clip}
Jie Liu, Yixiao Zhang, Jie-Neng Chen, Junfei Xiao, Yongyi Lu, Bennett A~Landman, Yixuan Yuan, Alan Yuille, Yucheng Tang, and Zongwei Zhou.
\newblock Clip-driven universal model for organ segmentation and tumor detection.
\newblock In \emph{Int. Conf. Comput. Vis.}, pages 21152--21164, 2023.

\bibitem[Liu et~al.(2019)Liu, Miao, Zhan, Wang, Gong, and Yu]{liu2019large}
Ziwei Liu, Zhongqi Miao, Xiaohang Zhan, Jiayun Wang, Boqing Gong, and Stella~X Yu.
\newblock Large-scale long-tailed recognition in an open world.
\newblock In \emph{IEEE Conf. Comput. Vis. Pattern Recog.}, pages 2537--2546, 2019.

\bibitem[Luo et~al.(2021)Luo, Chen, Song, and Wang]{luo2021semi}
Xiangde Luo, Jieneng Chen, Tao Song, and Guotai Wang.
\newblock Semi-supervised medical image segmentation through dual-task consistency.
\newblock In \emph{Proc. of the AAAI Conf. on Artificial Intelligence}, pages 8801--8809, 2021.

\bibitem[Lyu et~al.(2022)Lyu, Ye, Carlsen, Erleben, Darkner, and Yuen]{lyu2022pseudo}
Fei Lyu, Mang Ye, Jonathan~Frederik Carlsen, Kenny Erleben, Sune Darkner, and Pong~C Yuen.
\newblock Pseudo-label guided image synthesis for semi-supervised covid-19 pneumonia infection segmentation.
\newblock \emph{IEEE Trans. on Medical Imaging}, 42\penalty0 (3):\penalty0 797--809, 2022.

\bibitem[Peiris et~al.(2023)Peiris, Hayat, Chen, Egan, and Harandi]{peiris2023uncertainty}
Himashi Peiris, Munawar Hayat, Zhaolin Chen, Gary Egan, and Mehrtash Harandi.
\newblock Uncertainty-guided dual-views for semi-supervised volumetric medical image segmentation.
\newblock \emph{Nature Machine Intelligence}, 5\penalty0 (7):\penalty0 724--738, 2023.

\bibitem[Qi et~al.(2024)Qi, Wu, and Chan]{wu98gradient}
Wenbo Qi, Jiafei Wu, and SC Chan.
\newblock Gradient-aware for class-imbalanced semi-supervised medical image segmentation.
\newblock In \emph{Eur. Conf. Comput. Vis.}, 2024.

\bibitem[Qiao et~al.(2022)Qiao, Li, Song, Han, Gao, Tian, Liang, Li, Zhou, and Chen]{qiao2022semi}
Pengchong Qiao, Han Li, Guoli Song, Hu Han, Zhiqiang Gao, Yonghong Tian, Yongsheng Liang, Xi Li, S~Kevin Zhou, and Jie Chen.
\newblock Semi-supervised ct lesion segmentation using uncertainty-based data pairing and swapmix.
\newblock \emph{IEEE Trans. on Medical Imaging}, 2022.

\bibitem[Qiao et~al.(2018)Qiao, Shen, Zhang, Wang, and Yuille]{qiao2018deepcotraining}
Siyuan Qiao, Wei Shen, Zhishuai Zhang, Bo Wang, and Alan Yuille.
\newblock Deep co-training for semi-supervised image recognition.
\newblock In \emph{Eur. Conf. Comput. Vis.}, pages 135--152, 2018.

\bibitem[Radford et~al.(2021{\natexlab{a}})Radford, Kim, Hallacy, Ramesh, Goh, Agarwal, Sastry, Askell, Mishkin, Clark, et~al.]{radford2021clip}
Alec Radford, Jong~Wook Kim, Chris Hallacy, Aditya Ramesh, Gabriel Goh, Sandhini Agarwal, Girish Sastry, Amanda Askell, Pamela Mishkin, Jack Clark, et~al.
\newblock Learning transferable visual models from natural language supervision.
\newblock In \emph{Proc. of Intl. Conf. on Machine Learning}, pages 8748--8763, 2021{\natexlab{a}}.

\bibitem[Radford et~al.(2021{\natexlab{b}})Radford, Kim, Hallacy, Ramesh, Goh, Agarwal, Sastry, Askell, Mishkin, Clark, et~al.]{radford2021learning}
Alec Radford, Jong~Wook Kim, Chris Hallacy, Aditya Ramesh, Gabriel Goh, Sandhini Agarwal, Girish Sastry, Amanda Askell, Pamela Mishkin, Jack Clark, et~al.
\newblock Learning transferable visual models from natural language supervision.
\newblock In \emph{Proc. of Intl. Conf. on Machine Learning}, pages 8748--8763, 2021{\natexlab{b}}.

\bibitem[Shu et~al.(2019)Shu, Xie, Yi, Zhao, Zhou, Xu, and Meng]{shu2019meta}
Jun Shu, Qi Xie, Lixuan Yi, Qian Zhao, Sanping Zhou, Zongben Xu, and Deyu Meng.
\newblock Meta-weight-net: Learning an explicit mapping for sample weighting.
\newblock \emph{Adv. Neural Inform. Process. Syst.}, 32, 2019.

\bibitem[Tarvainen and Valpola(2017)]{tarvainen2017mean}
Antti Tarvainen and Harri Valpola.
\newblock Mean teachers are better role models: Weight-averaged consistency targets improve semi-supervised deep learning results.
\newblock \emph{Adv. Neural Inform. Process. Syst.}, 30, 2017.

\bibitem[Wang et~al.(2021)Wang, Zhai, Lasio, Zhang, Yi, Chen, Macvittie, Metaxas, Zhou, and Zhang]{wang2021semi}
Guotai Wang, Shuwei Zhai, Giovanni Lasio, Baoshe Zhang, Byong Yi, Shifeng Chen, Thomas~J Macvittie, Dimitris Metaxas, Jinghao Zhou, and Shaoting Zhang.
\newblock Semi-supervised segmentation of radiation-induced pulmonary fibrosis from lung ct scans with multi-scale guided dense attention.
\newblock \emph{IEEE Trans. on Medical Imaging}, 41\penalty0 (3):\penalty0 531--542, 2021.

\bibitem[Wang and Li(2023)]{wang2023dhc}
Haonan Wang and Xiaomeng Li.
\newblock {DHC}: Dual-debiased heterogeneous co-training framework for class-imbalanced semi-supervised medical image segmentation.
\newblock In \emph{Proc. of MICCAI}, pages 582--591, 2023.

\bibitem[Wang and Li(2024)]{wang2024towards}
Haonan Wang and Xiaomeng Li.
\newblock Towards generic semi-supervised framework for volumetric medical image segmentation.
\newblock \emph{Adv. Neural Inform. Process. Syst.}, 36, 2024.

\bibitem[Wang et~al.(2022)Wang, Wu, Lian, and Yu]{wang2022depl}
Xudong Wang, Zhirong Wu, Long Lian, and Stella~X Yu.
\newblock Debiased learning from naturally imbalanced pseudo-labels.
\newblock In \emph{IEEE Conf. Comput. Vis. Pattern Recog.}, pages 14647--14657, 2022.

\bibitem[Wang et~al.(2023)Wang, Zhang, Yu, and Xiao]{wang2023hunting}
Xiaoyang Wang, Bingfeng Zhang, Limin Yu, and Jimin Xiao.
\newblock Hunting sparsity: Density-guided contrastive learning for semi-supervised semantic segmentation.
\newblock In \emph{IEEE Conf. Comput. Vis. Pattern Recog.}, pages 3114--3123, 2023.

\bibitem[Wei et~al.(2021)Wei, Sohn, Mellina, Yuille, and Yang]{wei2021crest}
Chen Wei, Kihyuk Sohn, Clayton Mellina, Alan Yuille, and Fan Yang.
\newblock Crest: A class-rebalancing self-training framework for imbalanced semi-supervised learning.
\newblock In \emph{IEEE Conf. Comput. Vis. Pattern Recog.}, pages 10857--10866, 2021.

\bibitem[Xiang et~al.(2022)Xiang, Qiu, and Yang]{xiang2022fussnet}
Jinyi Xiang, Peng Qiu, and Yang Yang.
\newblock {FUSSNet}: Fusing two sources of uncertainty for semi-supervised medical image segmentation.
\newblock In \emph{Proc. of MICCAI}, pages 481--491, 2022.

\bibitem[Yu et~al.(2019)Yu, Wang, Li, Fu, and Heng]{yu2019uamt}
Lequan Yu, Shujun Wang, Xiaomeng Li, Chi-Wing Fu, and Pheng-Ann Heng.
\newblock Uncertainty-aware self-ensembling model for semi-supervised 3d left atrium segmentation.
\newblock In \emph{Proc. of MICCAI}, pages 605--613, 2019.

\bibitem[Zhang et~al.(2023{\natexlab{a}})Zhang, Xu, Usuyama, Xu, Bagga, Tinn, Preston, Rao, Wei, Valluri, et~al.]{zhang2023biomedclip}
Sheng Zhang, Yanbo Xu, Naoto Usuyama, Hanwen Xu, Jaspreet Bagga, Robert Tinn, Sam Preston, Rajesh Rao, Mu Wei, Naveen Valluri, et~al.
\newblock Biomedclip: a multimodal biomedical foundation model pretrained from fifteen million scientific image-text pairs.
\newblock \emph{arXiv preprint arXiv:2303.00915}, 2023{\natexlab{a}}.

\bibitem[Zhang et~al.(2023{\natexlab{b}})Zhang, Li, Chen, Yuille, Liu, and Zhou]{zhang2023continual}
Yixiao Zhang, Xinyi Li, Huimiao Chen, Alan~L Yuille, Yaoyao Liu, and Zongwei Zhou.
\newblock Continual learning for abdominal multi-organ and tumor segmentation.
\newblock In \emph{Proc. of MICCAI}, pages 35--45, 2023{\natexlab{b}}.

\end{thebibliography}
}

\end{document}